\documentclass[12pt,a4paper]{article}
\pdfoutput=1

\usepackage{soul}
\usepackage{url}
\usepackage[utf8]{inputenc}
\usepackage[small]{caption}
\usepackage{graphicx}
\usepackage{amsmath}
\usepackage{booktabs}
\usepackage{algorithm}
\usepackage{algorithmic}

\usepackage{subcaption}

\providecommand{\keywords}[1]{\textbf{\textit{Index terms---}} #1}

\newcommand{\embed}{\text{embed}}
\newcommand{\f}{\text{f}}
\newcommand{\itc}{\text{itc}}
\newcommand{\mfd}{\text{mfd}}
\newcommand{\nn}{\text{nn}}

\newcommand{\E}{\mathrm{E}}

\newcommand{\Latap}{\mathcal{L}_\mathrm{atap}}
\newcommand{\Latrl}{\mathcal{L}_\mathrm{atrl}}
\newcommand{\Lcerl}{\mathcal{L}_\mathrm{cerl}}
\newcommand{\Lfmrl}{\mathcal{L}_\mathrm{fmrl}}

\newcommand{\adamSingleFigureCS}[4]{%
  \begin{figure}
    \centering
    \includegraphics[width=#1\columnwidth]{#2}%
    \caption{#4}%
    \label{#3}
  \end{figure}
}

\newcommand{\adamIncludeFigure}[3]{
  \subcaptionbox{#2}{\includegraphics[width=#1\columnwidth]{#3}}
}

\newcommand{\adamIncludeFigureNC}[2]{
  \includegraphics[width=#1\columnwidth]{#2}
}

\newcommand{\adamIncludeFigureCS}[4]{
  \subcaptionbox{#3}[#2\columnwidth]{\includegraphics[width=#1\columnwidth]{#4}}
}

\newcommand{\figureVarVapComparison}{
  \noindent
  \begin{figure}
    \centering
    \captionsetup[sub]{labelformat=empty}
    \begin{tabular}{ccccccccc}
      \adamIncludeFigureNC{0.08}{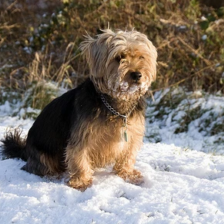}
      \adamIncludeFigureNC{0.08}{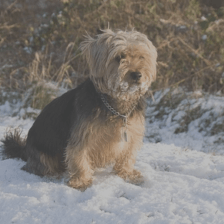}
      \adamIncludeFigureNC{0.08}{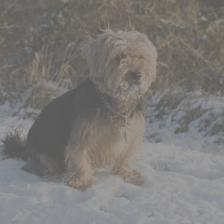}
      \adamIncludeFigureNC{0.08}{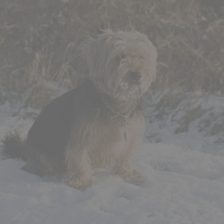}
      \adamIncludeFigureNC{0.08}{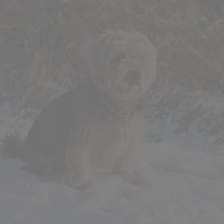}
      \adamIncludeFigureNC{0.08}{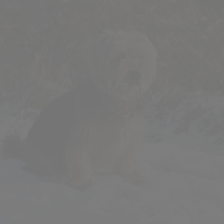}
      \adamIncludeFigureNC{0.08}{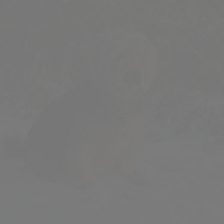}
      \adamIncludeFigureNC{0.08}{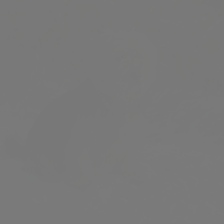}
      \adamIncludeFigureNC{0.08}{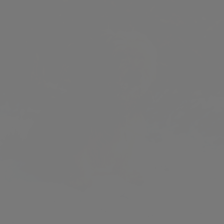}
      \\
      \adamIncludeFigureNC{0.08}{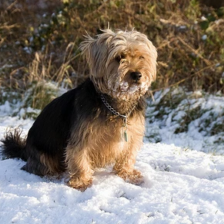}
      \adamIncludeFigureNC{0.08}{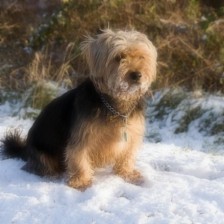}
      \adamIncludeFigureNC{0.08}{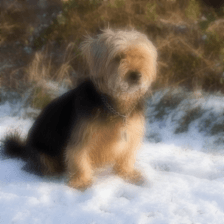}
      \adamIncludeFigureNC{0.08}{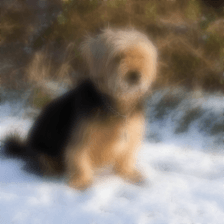}
      \adamIncludeFigureNC{0.08}{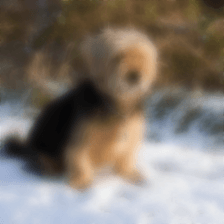}
      \adamIncludeFigureNC{0.08}{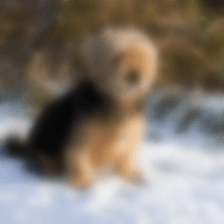}
      \adamIncludeFigureNC{0.08}{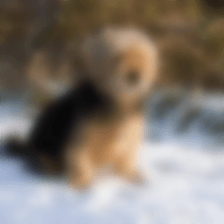}
      \adamIncludeFigureNC{0.08}{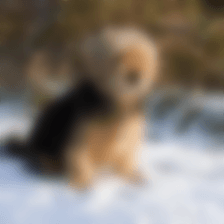}
      \adamIncludeFigureNC{0.08}{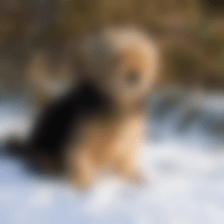}
    \end{tabular}
    \caption{The gradient impact comparison between variance and variance pooling during training. The first row shows the impact of variance while the second shows that of variance pooling. The visualization interval is 5000 steps of gradient backpropagation on the corresponding image.}%
    \label{fig:var_vap_comparison}
  \end{figure}
}

\newcommand{\figureEmbeddingExtractionArchitecture}{
  \begin{figure}
    \centering
    \begin{tabular}{cc}
      \adamIncludeFigure{0.45}{Embedding}{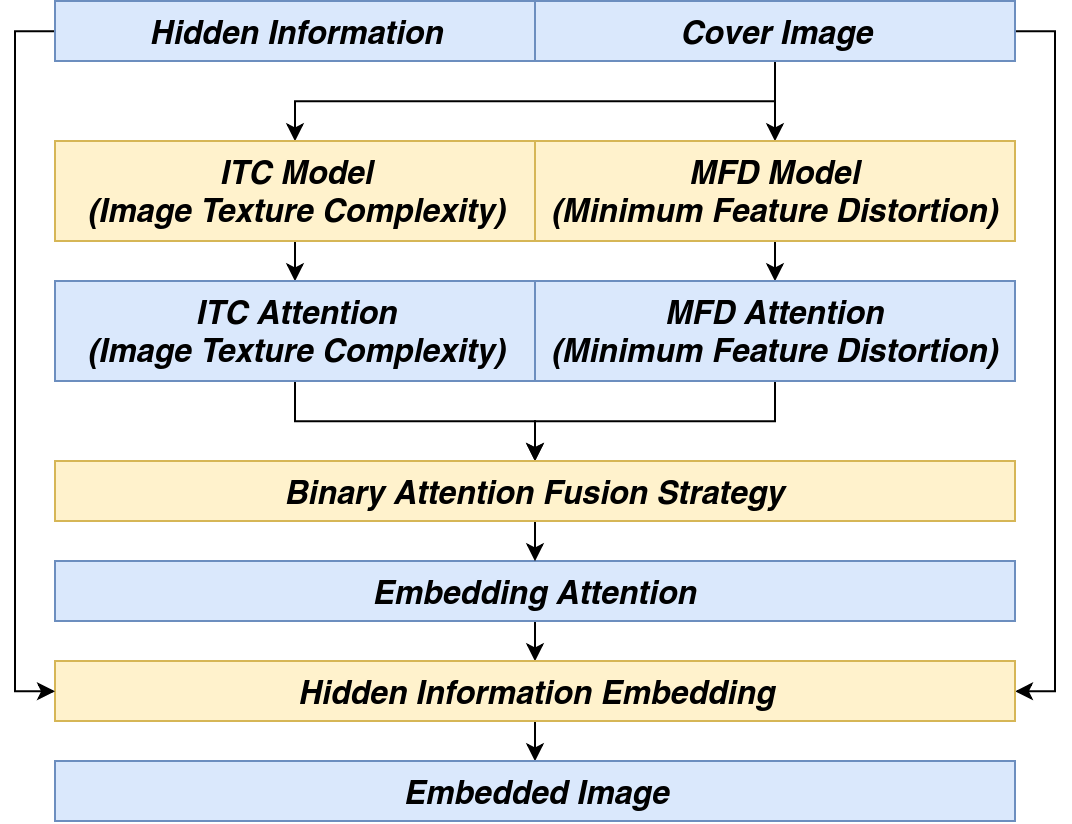}
      \adamIncludeFigure{0.45}{Extraction}{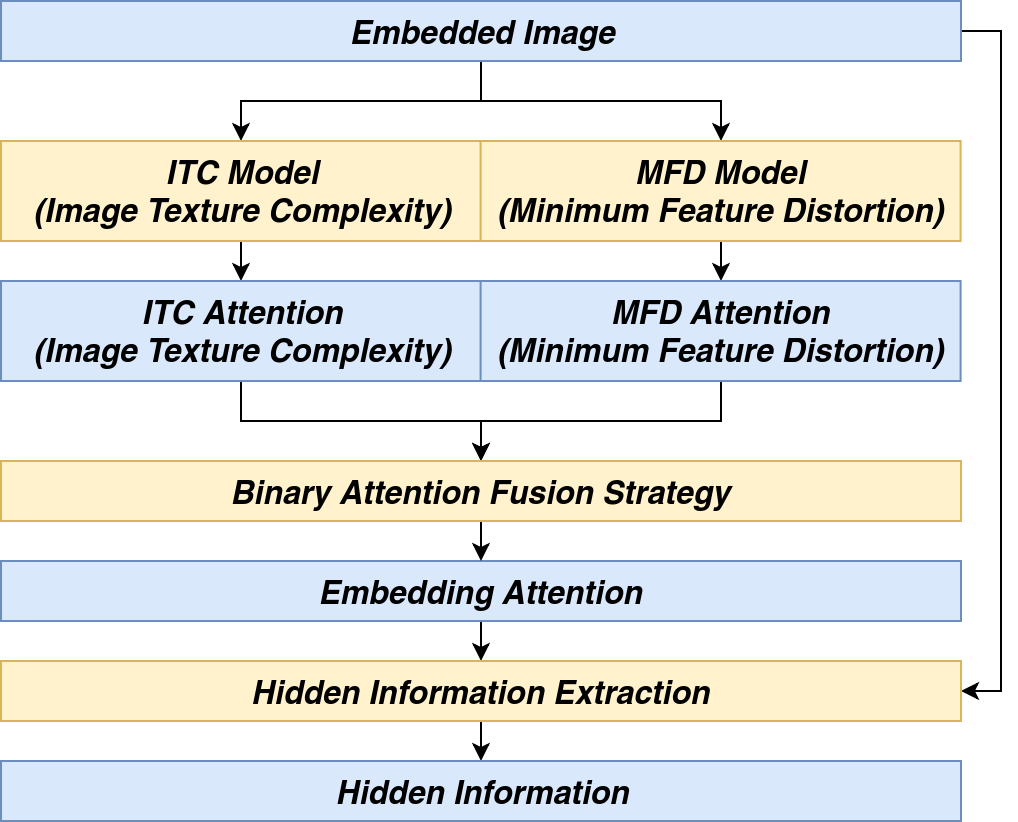}
    \end{tabular}
    \caption{The Embedding and Extraction Architecture}%
    \label{fig:embedding_extraction_architecture}
  \end{figure}
}

\newcommand{\figureTwoModelArchitectures}{
  \begin{figure}
    \centering
    \begin{tabular}{cc}
      \adamIncludeFigureCS{0.3}{0.4}{ITC Attention Model}{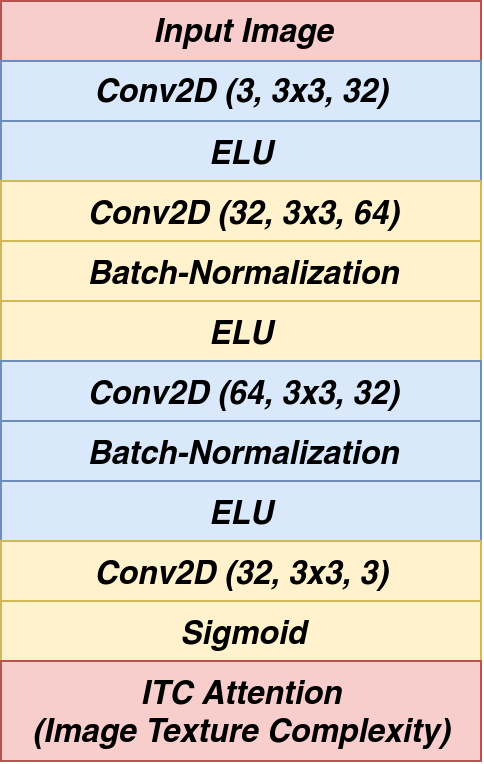}
      \adamIncludeFigureCS{0.5}{0.5}{MFD Attention Model}{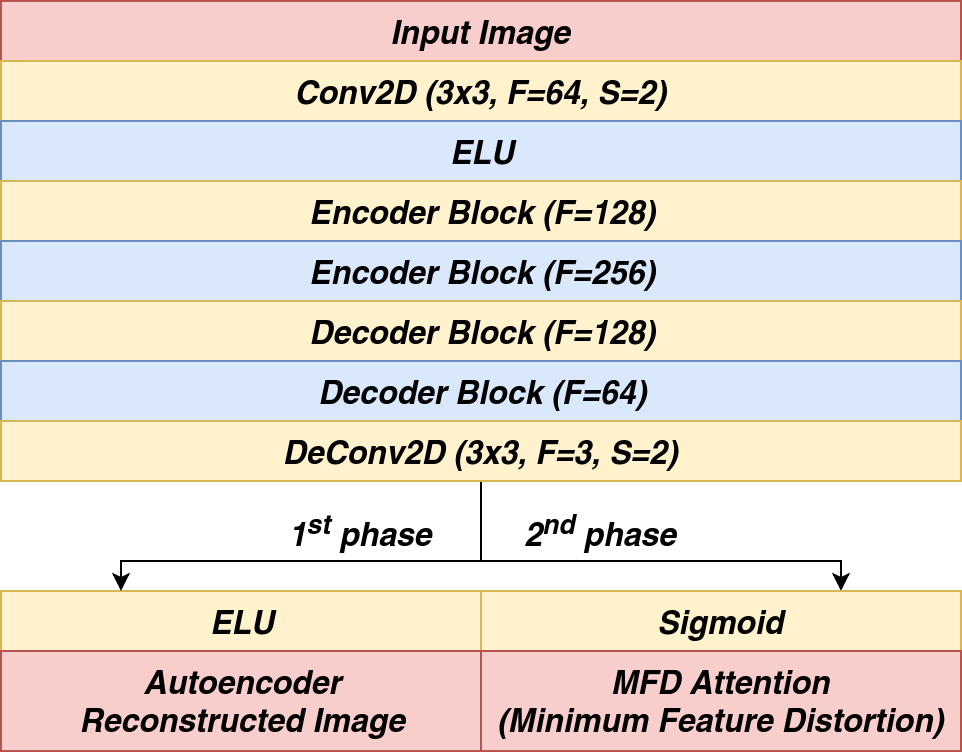}
    \end{tabular}
    \caption{Model Architectures}%
    \label{fig:two_model_architectures}
  \end{figure}
}

\newcommand{\figureItcAttentionEffect}{
  \begin{figure}
    \centering
    \begin{tabular}{ccc}
      \adamIncludeFigureCS{0.15}{0.25}{Original Image}{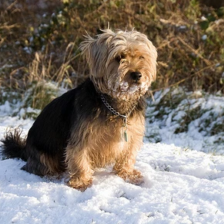}
      \adamIncludeFigureCS{0.15}{0.25}{ITC Attention}{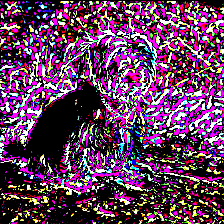}
      \adamIncludeFigureCS{0.15}{0.25}{Weighted Image}{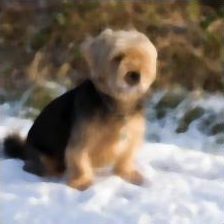}
      \\~\\
      \adamIncludeFigureCS{0.15}{0.25}{Original Image}{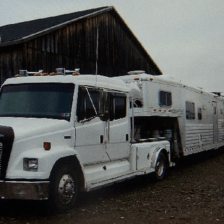}
      \adamIncludeFigureCS{0.15}{0.25}{ITC Attention}{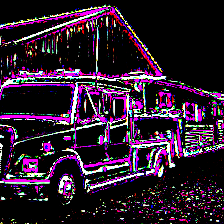}
      \adamIncludeFigureCS{0.15}{0.25}{Weighted Image}{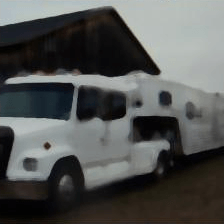}
    \end{tabular}
    \caption{The Effect of ITC Attention on Texture Complexity Reduction}%
    \label{fig:itc_attention_effect}
  \end{figure}
}

\newcommand{\figureItcAttentionFinetune}{
  \begin{figure}
    \centering
    \begin{tabular}{cccc}
      \adamIncludeFigure{0.15}{}{images/ImageSmoother/cover-00000000-0002.png}
      \adamIncludeFigure{0.15}{}{images/ITC/valid-attens-0000-0000-02.png}
      \adamIncludeFigure{0.15}{}{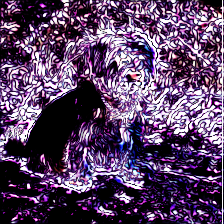}
      \adamIncludeFigure{0.15}{}{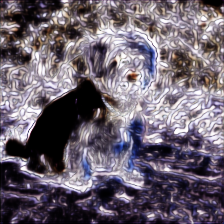}
      \\~\\
      \adamIncludeFigure{0.15}{}{images/ImageSmoother/cover-00000000-0023.png}
      \adamIncludeFigure{0.15}{}{images/ITC/valid-attens-0000-0000-23.png}
      \adamIncludeFigure{0.15}{}{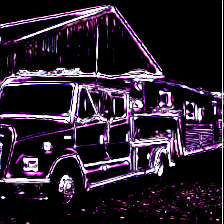}
      \adamIncludeFigure{0.15}{}{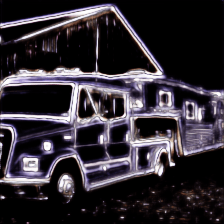}
    \end{tabular}
    \caption{ITC Attention After Finetune}%
    \label{fig:itc_attention_finetune}
    \vspace{\baselineskip}
    The first column shows the original image, the second column shows the ITC attention before any finetune, the third column shows the ITC attention after finetuning for minima fusion strategy, and the forth column shows the ITC attention after finetuning for mean fusion strategy.
  \end{figure}
}

\newcommand{\figureMfdAttentionEffect}{
  \begin{figure}
    \centering
    \begin{tabular}{ccc}
      \adamIncludeFigureCS{0.15}{0.3}{The Cover}{images/ImageSmoother/cover-00000000-0002.png}
      \adamIncludeFigureCS{0.15}{0.3}{MFD Attention}{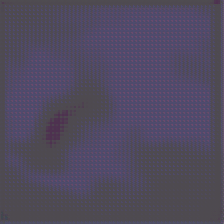}
      \adamIncludeFigureCS{0.15}{0.3}{The Embedded}{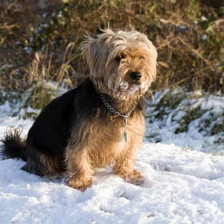}
      \\~\\
      \adamIncludeFigureCS{0.15}{0.3}{The Cover}{images/ImageSmoother/cover-00000000-0023.png}
      \adamIncludeFigureCS{0.15}{0.3}{MFD Attention}{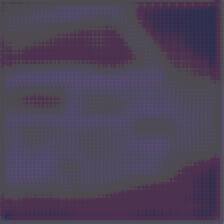}
      \adamIncludeFigureCS{0.15}{0.3}{The Embedded}{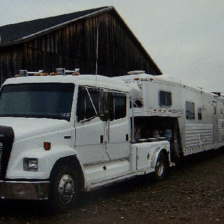}
    \end{tabular}
    \caption{The Visual Effect of MFD Attention on Embedding with Random Noise}%
    \label{fig:mfd_attention_effect}
  \end{figure}
}

\newcommand{\figureMfdEncoderDecoderBlock}{
  \begin{figure}
    \centering
    \begin{tabular}{cc}
      \adamIncludeFigure{0.35}{Encoder}{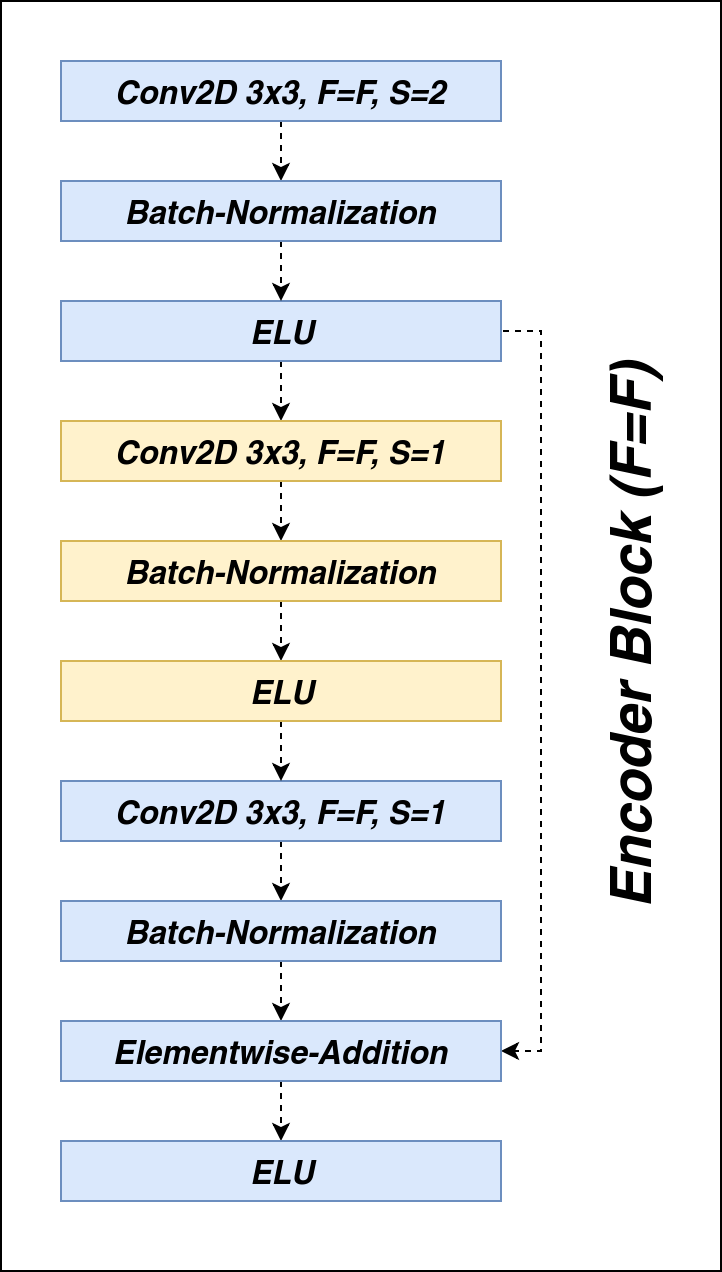}
      \adamIncludeFigure{0.35}{Decoder}{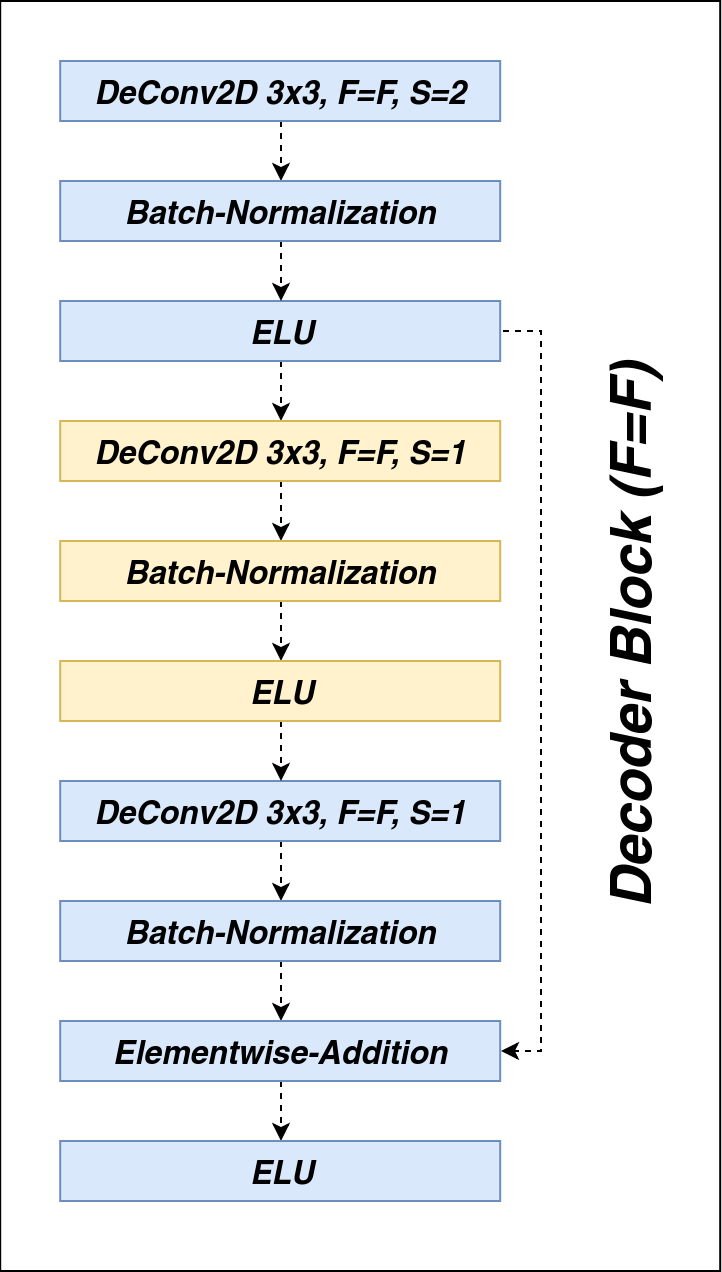}
    \end{tabular}
    \caption{The Encoder and Decoder Block of the MFD Attention Model}%
    \label{fig:mfd_encoder_decoder_block}
  \end{figure}
}

\newcommand{\figureImageSmoothingComparison}{
  \begin{figure}
    \centering
    \begin{tabular}{cccc}
      \adamIncludeFigureCS{0.15}{0.2}{Original}{images/ImageSmoother/cover-00000000-0002.png}
      \adamIncludeFigureCS{0.15}{0.2}{Average} {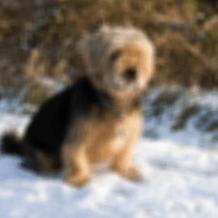}
      \adamIncludeFigureCS{0.15}{0.2}{Gaussian}{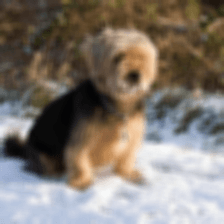}
      \adamIncludeFigureCS{0.15}{0.2}{Median}  {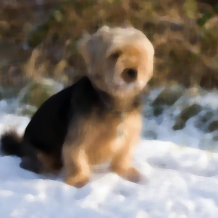}
      \\~\\
      \adamIncludeFigureCS{0.15}{0.2}{Original}{images/ImageSmoother/cover-00000000-0023.png}
      \adamIncludeFigureCS{0.15}{0.2}{Average} {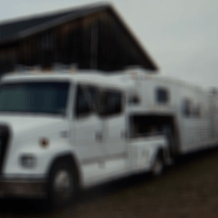}
      \adamIncludeFigureCS{0.15}{0.2}{Gaussian}{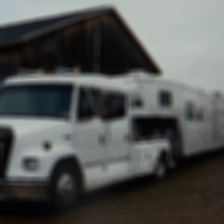}
      \adamIncludeFigureCS{0.15}{0.2}{Median}  {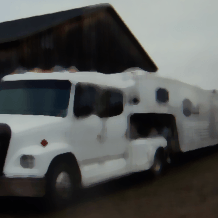}
    \end{tabular}
    \caption{Image Smoothing Effect Comparison}%
    \label{fig:image_smoothing_comparison}
  \end{figure}
}

\newcommand{\figureSteganographyResultMeanLSMOne}{
  \begin{figure}
    \centering
    \begin{tabular}{ccc}
      \adamIncludeFigureCS{0.15}{0.3}{The Cover}      {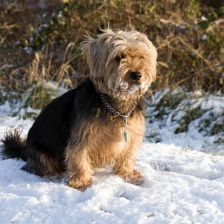}
      \adamIncludeFigureCS{0.15}{0.3}{Fused Attention}{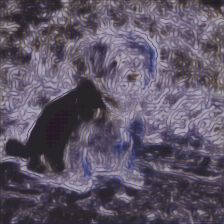}
      \adamIncludeFigureCS{0.15}{0.3}{The Embedded}   {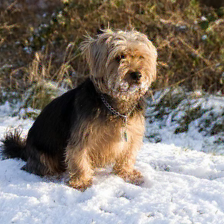}
      \\~\\
      \adamIncludeFigureCS{0.15}{0.3}{The Cover}      {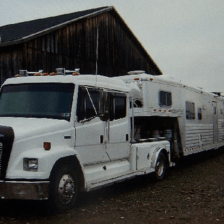}
      \adamIncludeFigureCS{0.15}{0.3}{Fused Attention}{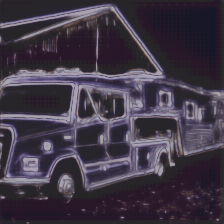}
      \adamIncludeFigureCS{0.15}{0.3}{The Embedded}   {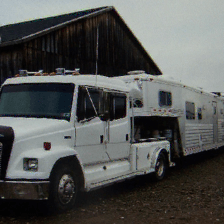}
    \end{tabular}
    \caption{Steganography using Mean Fusion with 1-bit LSM}%
    \label{fig:steganography_result_mean_lsm_1}
  \end{figure}
}

\newcommand{\figureROCCurves}{
  \begin{figure}
    \centering
    \begin{tabular}{cc}
      \adamIncludeFigure{0.48}{StegExpose}      {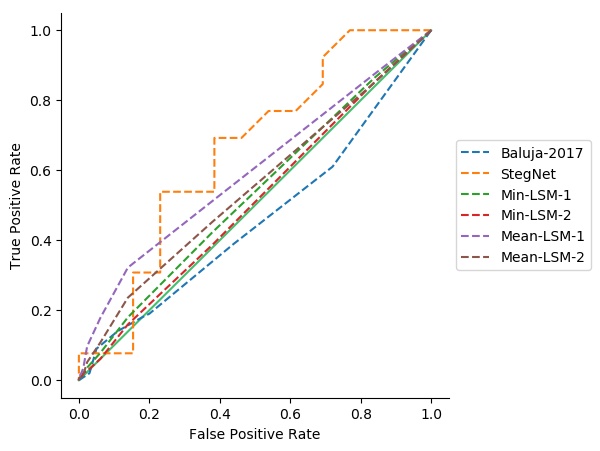}%
      \adamIncludeFigure{0.48}{SPAM Features}   {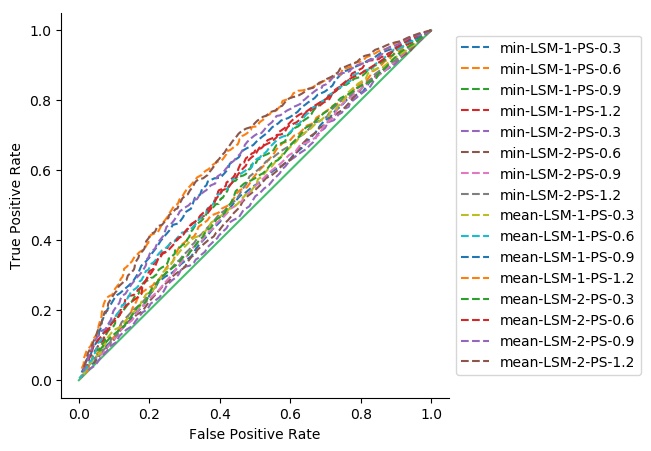}%
      \\~\\
      \adamIncludeFigure{0.48}{SRM Features}    {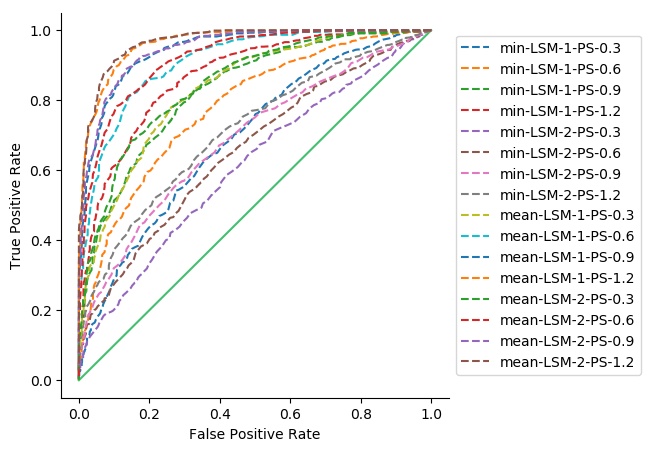}
      \adamIncludeFigure{0.48}{YedroudjNet}     {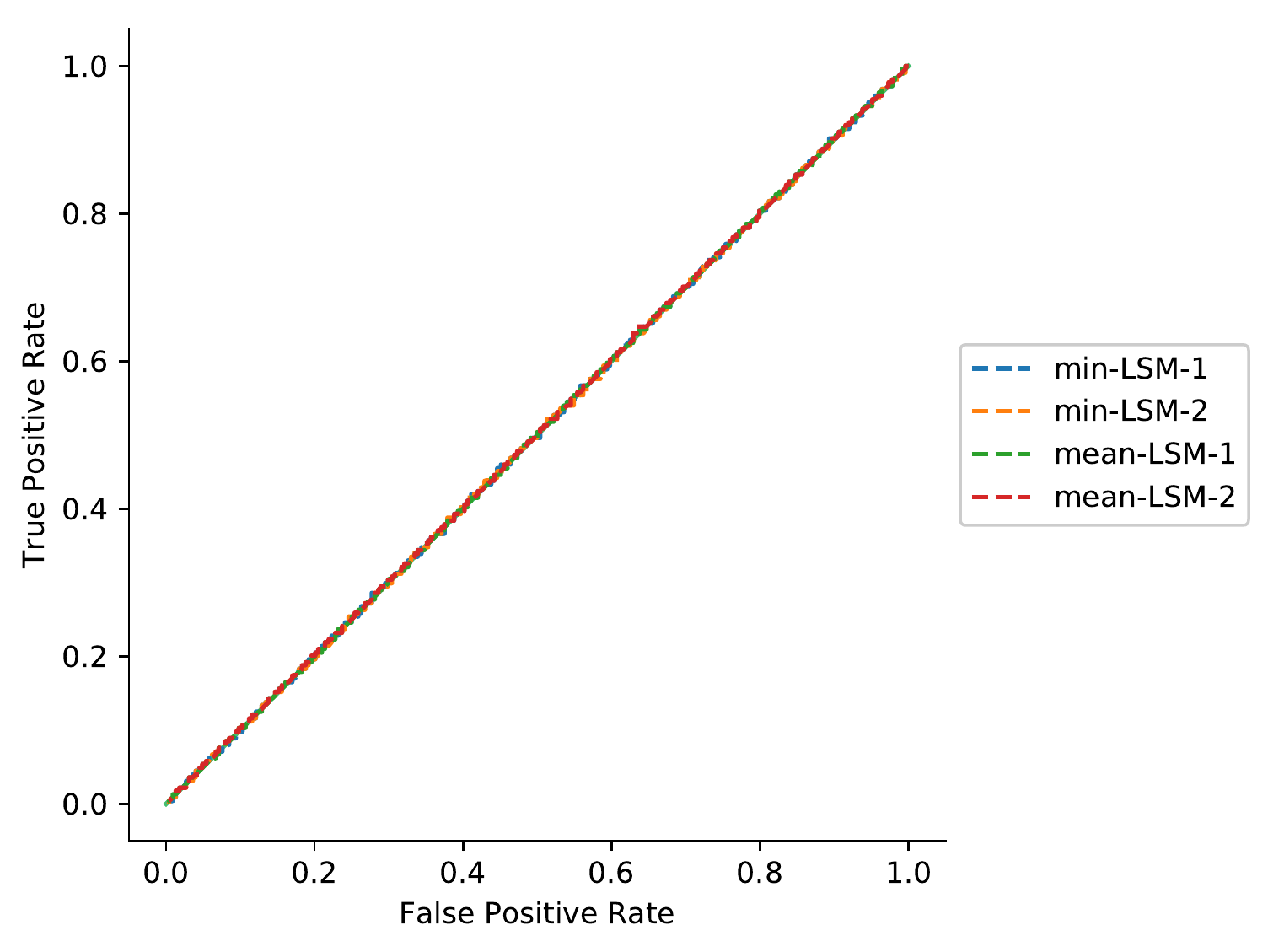}%
    \end{tabular}
    \caption{ROC Curves: Steganalysis with StegExpose, SPAM Features, SRM Features and Yedroudj-Net}%
    \label{fig:roc_curves}
  \end{figure}
}

\newcommand{\figureBASNFeatureDistortionRate}{
  \adamSingleFigureCS{0.6}{%
    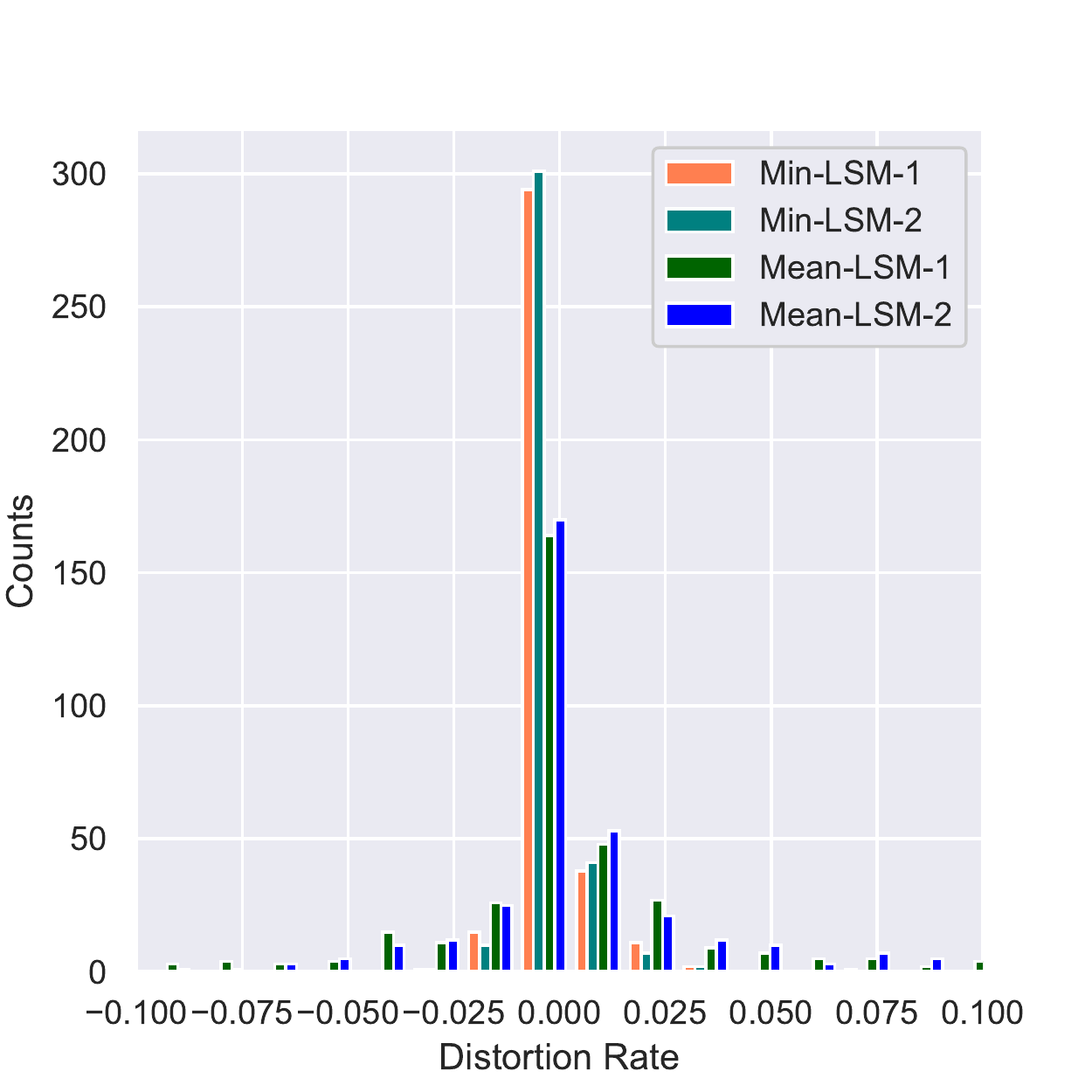%
  }{%
    fig:basn_feature_distortion_rate%
  }{
    ResNet-18 Classification Feature Distortion Rate
  }
}

\newcommand{\figureSoftAreaPenalty}{
  \begin{figure}
    \centering
    \begin{tabular}{cc}
      \adamIncludeFigure{0.3}{ITC Area Penalty}{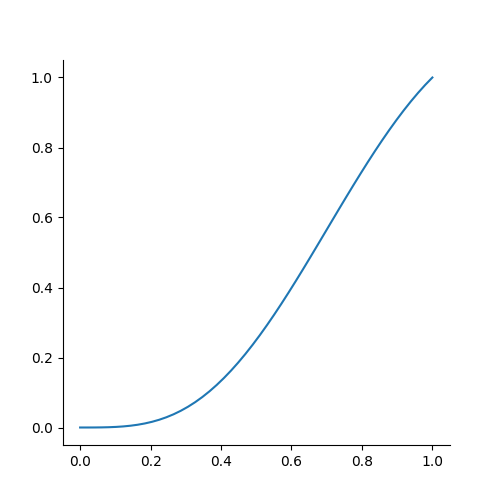}
      \adamIncludeFigure{0.3}{MFD Area Penalty}{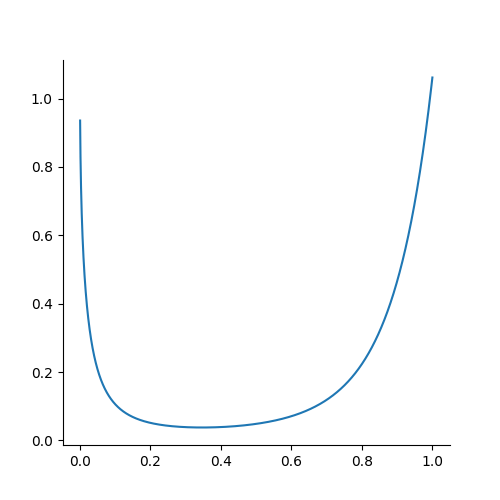}
    \end{tabular}
    \caption{Soft Area Penalties}%
    \label{fig:soft_area_penalty}
  \end{figure}
}

\newcommand{\figureFinetuneFirstPhase}{
  \begin{figure}
    \centering
    \includegraphics[width=0.9\columnwidth]{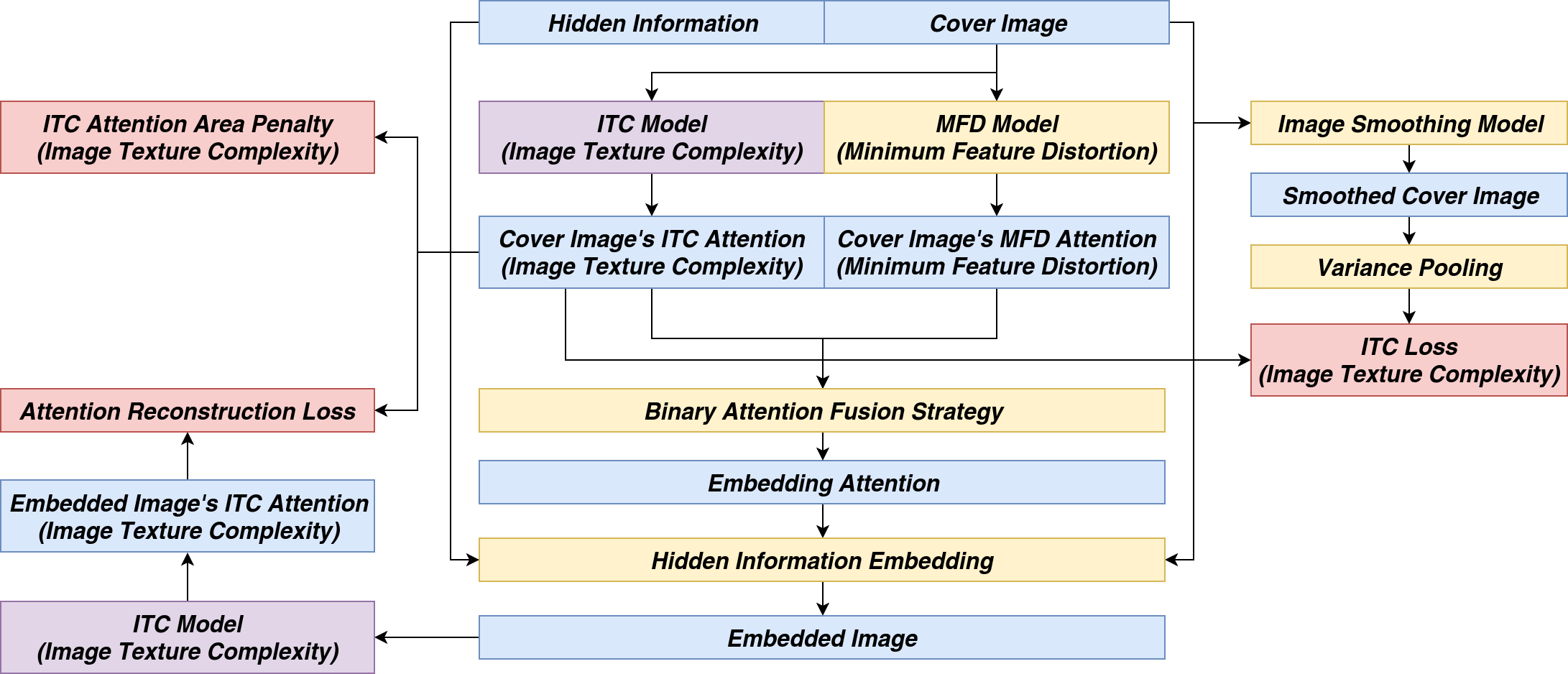}
    \caption{The \(1^{st}\) Phase Finetune Pipeline}%
    \label{fig:finetune_first_phase}
  \end{figure}
}

\newcommand{\figureMFDModelOverall}{
  \begin{figure}
    \centering
    \includegraphics[width=0.8\columnwidth]{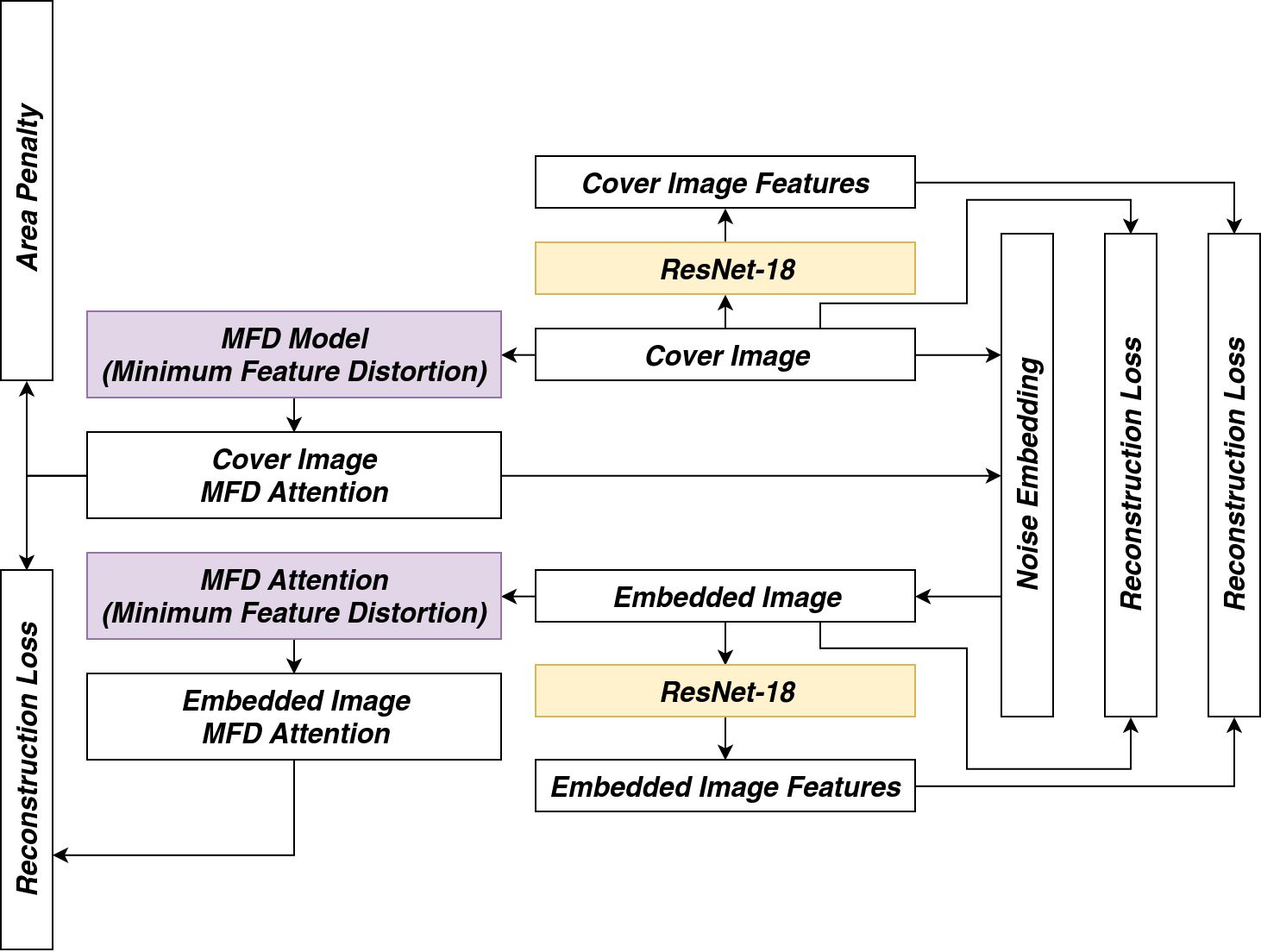}
    \caption{MFD Attention Mechanism Training Pipeline}%
    \label{fig:mfd_model_overall}
  \end{figure}
}

\author{%
  Yang Yang
  \thanks{ORCID: 0000-0002-6627-2987}
  \thanks{E-Mail: GetbetterABC at yeah.net}
  \thanks{Email: adamcavendish at shu.edu.cn}
}

\title{BASN --- Learning Steganography with Binary Attention Mechanism}

\begin{document}
\maketitle

\maketitle
\begin{abstract}
  Secret information sharing through image carrier has aroused much research attention in recent years with images' growing domination on the Internet and mobile applications. However, with the booming trend of convolutional neural networks, image steganography is facing a more significant challenge from neural-network-automated tasks. To improve the security of image steganography and minimize task result distortion, models must maintain the feature maps generated by task-specific networks being irrelative to any hidden information embedded in the carrier. This paper introduces a binary attention mechanism into image steganography to help alleviate the security issue, and in the meanwhile, increase embedding payload capacity. The experimental results show that our method has the advantage of high payload capacity with little feature map distortion and still resist detection by state-of-the-art image steganalysis algorithms.~\footnote{Source code will be published at: \url{https://github.com/adamcavendish/BASN-Learning-Steganography-with-Binary-Attention-Mechanism}}%
\end{abstract}

\section{Introduction}

\begin{figure}
  \centering
  \includegraphics[width=0.9\columnwidth]{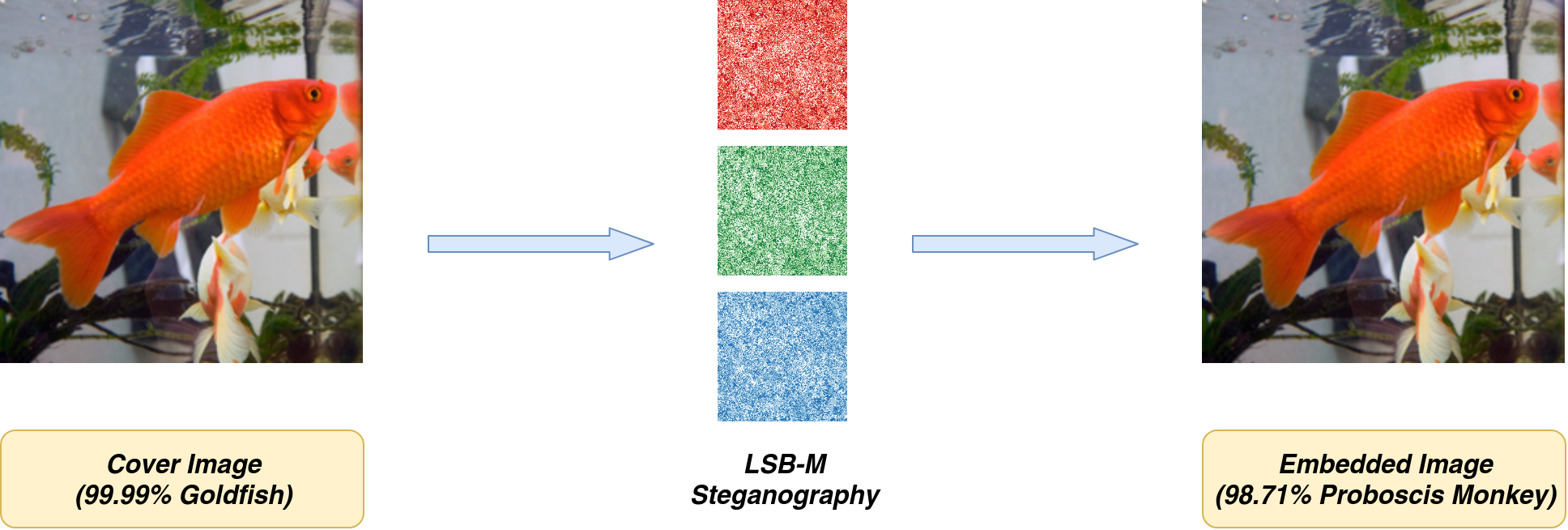}%
  \caption{LSB-Matching Embedded Image Misclassification}%
  \label{fig:steganography_distortion}
  \vspace{\baselineskip}
  The cover image and embedded image both use ImageNet pretrained ResNet-18~\cite{ResNet} network for classification. The percentage before the predicted class label represents network's confidence in prediction. The red, green and blue noisy images in the center represent the altered pixel locations in corresponding channels during steganography. There're only three kinds of colors within these images where white stands for no modification, the lighter one stands for a +1 modification and the darker one stands for a -1 modification.
\end{figure}

Image steganography aims at delivering a modified cover image to secretly transfer hidden information inside with little awareness of the third-party supervision. On the other side, steganalysis algorithms are developed to find out whether an image is embedded with hidden information or not, and therefore, resisting steganalysis detection is one of the major indicators of steganography security. In the meanwhile, with the booming trend of convolutional neural networks, a massive amount of neural-network-automated tasks are coming into industrial practices like image auto-labeling through object detection~\cite{Fast_R_CNN,YOLO} and classification~\cite{ResNet,InceptionV4}, face recognition~\cite{FaceNet}, pedestrian re-identification~\cite{CamStylePedestrian} and etc. Images steganography is now facing a more significant challenge from these automated tasks, whose embedding distortion might influcence the task result in a great manner and irresistibly lead to suspicion. Figure~\ref{fig:steganography_distortion} is an example that LSB-Matching~\cite{LSBRevisited} steganography completely alters the image classification result from goldfish to proboscis monkey. Under such circumstances, a steganography model even with outstanding invisibility to steganalysis methods still cannot be called secure where the spurious label might re-arouse suspicion and finally, all efforts are made in vain.

\subsection{Related Works}

Most previous steganography models focus on resisting steganalysis algorithms or raising embedding payload capacity. BPCS~\cite{BPCS2002,BPCS2015} and PVD~\cite{PVD,PVD_LSB,PVD_Mod} uses adaptive embedding based on local complexty to improve embedding visual quality. HuGO~\cite{HUGO} and S-UNIWARD~\cite{S_UNIWARD} resist steganalysis by minimizing a suitably defined distortion function. Hu~\cite{GANStego} adopts deep convolutional generative adversarial network to achieve steganography without embedding. Wu~\cite{StegNet} and Baluja~\cite{HIPS} achieve a vast payload capacity by focusing on image-into-image steganography.

\subsection{Contributions of this work}

In this paper, we propose a Binary Attention Steganography Network (abbreviated as \textbf{BASN}) architecture to achieve a relatively high payload capacity (2-3 bpp) with minimal distortion to other neural-network-automated tasks. It utilizes convolutional neural networks with two attention mechanisms, which minimizes embedding distortion to the human visual system and neural network feature maps respectively. Additionally, multiple attention fusion strategies are suggested to balance payload capacity with security, and a fine-tuning mechanism are put forward to improve the hidden information extraction accuracy.

\section{Binary Attention Mechanism}

Binary attention mechanism involves two attention models including image texture complexity (ITC) attention model and minimizing feature distortion (MFD) attention model. ITC model mainly focuses on deceiving the human visual system from noticing the differences out of altered pixels. MFD model minimizes the high-level features extracted between clean and embedded images so that neural networks will not give out diverge results. The attention mechanism in both models serve as a hint for steganography showing where to embed and how much information the corresponding pixel might tolerate.

The embedding and extraction overall architecture are shown in Figure~\ref{fig:embedding_extraction_architecture}. After two attentions are found with the binary attention mechanism, we may adopt several fusion strategies to create the final attention used for embedding and extraction.

\figureEmbeddingExtractionArchitecture%

\subsection{Evaluation of Image Texture Complexity}

To evaluate an image's texture complexity, variance is adapted in most approaches. However, using variance as the evaluation mechanism enforces very strong pixel dependencies. In other words, every pixel is correlated to all other pixels in the image.

We propose variance pooling evaluation mechanism to relax cross-pixel dependencies (See Equation~\ref{eq:var_pool2d}). Variance pooling applies on patches but not the whole image to restrict the influence of pixel value alterations within the corresponding patches. Especially in the case of training when optimizing local textures to reduce its complexity, pixels within the current area should be most frequently changed while far distant ones are intended to be reserved for keeping the overall image contrast, brightness and visual patterns untouched.

\begin{equation}
  \mathrm{VarPool2d}(X_{i,j}) %
    = \E_{k_i} \left( \E_{k_j} \left( X_{i+k_i,j+k_j}^2 \right) \right) %
    - \E_{k_i} \left( \E_{k_j} {\left( X_{i+k_i,j+k_j} \right)}^2 \right)
  \label{eq:var_pool2d}
\end{equation}

\begin{equation}%
  k_i \in \left[ -\frac{n}{2}, \frac{n}{2} \right],~ k_j \in \left[ -\frac{n}{2}, \frac{n}{2} \right]
  \label{eq:var_pool2d_cond}
\end{equation}

In Equation~\ref{eq:var_pool2d}, \(X\) is a 2-dimensional random variable which can be either an image or a feature map and \(i, j\) are the indices of each dimension. Operator \(\mathrm{E}(\cdot)\) calculates the expectation of the random variable. VarPool2d applies similar kernel mechanism as other 2-dimensional pooling or convolution operations and \(k_i, k_j\) indicates the kernel indices of each dimension.

To further show the impact of gradients updating between variance and variance pooling during backpropagation, we applied the gradients backpropagated directly to the image to visualize how gradients influences the image itself during training (See Equation~\ref{eq:varvap_comp_loss_variance},\ref{eq:varvap_comp_loss_varpool2d} for training loss and Figure~\ref{fig:var_vap_comparison} for the impact comparison).

\begin{align}
    \mathcal{L}_\mathrm{Variance}  &= \mathrm{Variance}(X) \label{eq:varvap_comp_loss_variance} \\
    \mathcal{L}_\mathrm{VarPool2d} &= \E \left( \mathrm{VarPool2d}_{n=7} \left( X \right) \right) \label{eq:varvap_comp_loss_varpool2d}
\end{align}

\figureVarVapComparison%

\subsection{ITC Attention Model}

ITC (Image Texture Complexity) attention model aims to embed information without being noticed by the human visual system, or in other words, making just noticeable difference (JND) to cover images to ensure the largest embedding payload capacity~\cite{JND}. In texture-rich areas, it is possible to alter pixels to carry hidden information without being noticed. Finding the ITC attention means finding the positions of the image pixels and their corresponding capacity that tolerate mutations.

Here we introduce two concepts:
\begin{enumerate}
  \item A hyper-parameter \( \theta \) representing the ideal embedding payload capacity that the input image might achieve.
  \item An ideal texture-free image \(C_{\theta}\) corresponding to the input image that is visually similar but with the lowest texture complexity possible regarding the restriction of at most \( \theta \) changes.
\end{enumerate}

With the help of these concepts, we can formulate the aim of ITC attention model as:

For each cover image \(C\), ITC model \(f_{\itc}\) needs to find an attention \(A_{\itc} = f_{\itc}(C)\) to minimize the texture complexity evaluation function \(V_{\itc}\):

\begin{align}
  \text{minimize}   \quad & V_{\itc}(A_{\itc} \cdot C_{\theta} + (1 - A_{\itc}) \cdot C)
  \label{eq:itc_minimize}                                                                         \\
  \text{subject to} \quad & \frac{1}{N} \sum_{i}^{N} A_{\itc} \leq \theta \label{eq:itc_subject_to}
\end{align}

The \( \theta \) in Equation~\ref{eq:itc_subject_to} is used as an upper bound to limit down the attention area size. If trained without it, model \(f_{\itc}\) is free to output all-ones matrix \(A_{\itc}\) to acquire an optimal texture-free image. It is well-known that an image with the least amount of texture is a solid color image, which does not help find the correct texture-rich areas.

In actual training process, the detailed model architecture is shown in Figure~\ref{fig:two_model_architectures} and two parts of the equation are slightly modified to ensure better training results. First, the ideal texture-free image \(C_{\theta}\) in Equation~\ref{eq:itc_minimize} does not indeed exist but is available through approximation nonetheless. In this paper median pooling with a kernel size of 7 is used to simulate the ideal texture-free image. It helps eliminate detailed textures within patches without touching object boundaries (See Figure~\ref{fig:image_smoothing_comparison} for comparison among different smoothing techniques). Second, we adopt soft bound limits in place of hard upper bound in forms of Equation~\ref{eq:itc_area_penalty} (visualized in Figure~\ref{fig:soft_area_penalty}). Soft limits help generate smoothed gradients and provide optimizing directions.

\figureImageSmoothingComparison%

\begin{equation}
  \text{Area-Penalty}_\itc = {\E(A_{\itc})}^{3-2 \cdot \E(A_{\itc})}
  \label{eq:itc_area_penalty}
\end{equation}

\figureItcAttentionEffect%

The overall loss on training ITC attention model is listed in Equation~\ref{eq:itc_var_loss},\ref{eq:itc_overall_loss}, and Figure~\ref{fig:itc_attention_effect} shows the effect of ITC attention on image texture complexity reduction. The attention area reaches 21.2\% on average, and the weighted images gain an average of 86.3\% texture reduction in the validation dataset.

\begin{equation}
  \mathrm{VarLoss} = \E \left( \mathrm{VarPool2d} \left( A_{\itc} \cdot C_{\theta} + (1 - A_{\itc}) \cdot C \right) \right)
  \label{eq:itc_var_loss}
\end{equation}

\begin{equation}
  \mathrm{Loss}_\itc = \lambda \cdot \text{VarLoss} + (1 - \lambda) \cdot \mathrm{Area-Penalty}_\itc
  \label{eq:itc_overall_loss}
\end{equation}

\subsection{MFD Attention Model}
MFD (Minimizing Feature Distortion) attention model aims to embed information with least impact on neural network extracted features. Its attention also indicates the position of image pixels and their corresponding capacity that tolerate mutations.

For each cover image \(C\), MFD model \(f_{\mfd}\) needs to find an attention \(A_{\mfd} = f_{\mfd}(C)\) that minimizes the distance between cover image features \(f_{\nn}(C)\) and embedded image features \(f_{\nn}(S)\) after embedding information into cover image according to its attention.

\begin{equation}
  S = f_{\embed}(C, A_{\mfd})
\end{equation}

\begin{align}
  \text{minimize}   \quad & \Lfmrl(f_{\nn}(C), f_{\nn}(S)) \\
  \text{subject to} \quad & \alpha \leq \frac{1}{N} \sum_{i}^{N} A_{\mfd} \leq \beta
\end{align}

Here, \(C\) stands for the cover image and \(S\) stands for the corresponding embedded image. \(\Lfmrl(\cdot)\) is the feature map reconstruction loss and \( \alpha, \beta \) are thresholds limiting the area of attention map acting the same role as \( \theta \) in the ITC attention model.

\figureTwoModelArchitectures%

\figureMFDModelOverall%
 
\figureMfdEncoderDecoderBlock%

The actual ways of training the MFD attention model is split into 2 phases (See Figure~\ref{fig:two_model_architectures}). The first training phase aims to initialize the weights of encoder blocks using the left path shown in Figure~\ref{fig:two_model_architectures} as an autoencoder. In the second training phase, all the weights of decoder blocks are reset and takes the right path to generate MFD attentions. The encoder and decoder block architectures are shown in Figure~\ref{fig:mfd_encoder_decoder_block}.

The overall training pipeline in the second phase is shown in Figure~\ref{fig:mfd_model_overall}. The weights of two MFD blocks colored in purple are shared while the weights of two task specific neural network blocks colored in yellow are frozen. In the training process, task specific neural network works only as a feature extractor and therefore it can be simply extended to multiple tasks by reshaping and concatenating feature maps together. Here we adopt ResNet-18~\cite{ResNet} as an example for minimizing embedding distortion to the classification task.

The overall loss on training MFD attention model (phase 2) is listed in Equation~\ref{eq:mfd_overall_loss}.
The \( \Lfmrl \) (Feature Map Reconstruction Loss) uses \(L_2\) loss to reconstruct between cover image extracted feature maps and embedded ones. The \( \Lcerl \) (Cover Embedded image Reconstruction Loss) and \( \Latrl \) (Attention Reconstruction Loss) uses \(L_1\) loss to reconstruct between the cover images and the embedded images and their corresponding attentions. The \( \Latap \) (ATtention Area Penalty) also applies soft bound limit in forms of Equation~\ref{eq:mfd_area_penalty} (visualized in Figure~\ref{fig:soft_area_penalty}). The visual effect of MFD attention embedding with random noise is shown in Figure~\ref{fig:mfd_attention_effect}.

\begin{equation}
  \mathrm{Loss}_\mfd = \Lfmrl + \Lcerl + \Latrl + \Latap
  \label{eq:mfd_overall_loss}
\end{equation}

\begin{equation}
  \text{Area-Penalty}_\mfd = \frac{1}{2} \cdot {(1.1 \cdot \E(A_{\mfd}))}^{8 \cdot \E(A_{\mfd}) - 0.1}
  \label{eq:mfd_area_penalty}
\end{equation}

\figureSoftAreaPenalty%

\figureMfdAttentionEffect%

\section{Fusion Strategies, Finetune Process and Inference Techniques}

The fusion strategies help merge ITC and MFD attention models into one attention model, and thus they are substantial to be consistent and stable. In this paper, two fusion strategies being minima fusion and mean fusion are put forth as Equation~\ref{eq:min_fusion} and~\ref{eq:mean_fusion}. Minima fusion strategy aims to improve security while mean fusion strategy generates more payload capacity for embedding.

\begin{equation}
  A_{\f} = \min(A_{\itc}, A_{\mfd})
  \label{eq:min_fusion}
\end{equation}

\begin{equation}
  A_{\f} = \frac{1}{2}(A_{\itc}, A_{\mfd})
  \label{eq:mean_fusion}
\end{equation}

After a fusion strategy is applied, finetuning process is required to improve attention reconstruction on embedded images. The finetune process is split into two phases. In the first phase, the ITC model is finetuned as Figure~\ref{fig:finetune_first_phase}. The two ITC model colored in purple shares the same network weights and the MFD model weights are freezed. Besides from the image texture complexity loss (Equation~\ref{eq:itc_var_loss}) and the ITC area penalty (Equation~\ref{eq:itc_area_penalty}), the loss additionally involves an attention reconstruction loss using \(L_1\) loss similar to \( \Latrl \) in Equation~\ref{eq:mfd_overall_loss}. In the second phase, the new ITC model is freezed, and the MFD model is finetuned using its original loss (Equation~\ref{eq:mfd_overall_loss}).

\figureFinetuneFirstPhase%

The ITC model, after finetune, appears to be more interested in the texture-complex areas while ignores the areas that might introduce noises into the attention (See Figure~\ref{fig:itc_attention_finetune}).

\figureItcAttentionFinetune%

When using the model for inference after finetuning, two extra techniques are proposed to strengthen steganography security. The first technique is named \textit{Least Significant Masking (LSM)} which masks the lowest several bits of the attention during embedding. After the hidden information is embedded, the masked bits are restored to the original data to disturb the steganalysis methods. The second technique is called \textit{Permutative Straddling}, which sacrifices some payload capacity to straddle between hidden bits and cover bits~\cite{F5Stego}. It is achieved by scattering the effective payload bit locations across the overall embedded locations using a random seed. The overall hidden bits are further re-arranged sequentially in the effective payload bit locations. The random seed is required to restore the hidden data.

\section{Experiments}

\subsection{Experiments Configurations}

To demonstrate the effectiveness of our model, we conducted experiments on ImageNet dataset~\cite{ImageNet}. Specially, ILSVRC2012 dataset with 1,281,167 images is used for training and 50,000 for testing. Our work is trained on one NVidia GTX1080 GPU and we adopt a batch size of 32 for all models. Optimizers and learning rate setup for ITC model, MFD model \(1^{st}\) phase and MFD model \(2^{nd}\) phase are Adam optimizer~\cite{Adam} with 0.01, Nesterov momentum optimizer~\cite{Nesterov} with 1e-5 and Adam optimizer with 0.01 respectively.

All the validation processes use the compressed version of \textit{The Complete Works of William Shakespeare}~\cite{Shakespeare} provided by Project Gutenberg~\cite{Gutenberg}. It is downloaded here at~\cite{GutenbergShakespeare}.

The error rate uses BSER (Bit Steganography Error Rate) shown in Equation~\ref{eq:BSER}.

\begin{equation}
  \text{BSER} = \frac{\text{Number of redundant bits or missing bits}}{\text{Number of hidden information bits}} \times 100\%
  \label{eq:BSER}
\end{equation}

\subsection{Different Embedding Strategies Comparison}

Table~\ref{tab:diff_strategies_comparison} presents a performance comparison among different fusion strategies and different inference techniques. These techniques offer several ways to trade off between error rate and payload capacity. With \textit{Permutative Straddling}, it is further possible to precisely handle the payload capacity during transmission.

\begin{table}
  \centering
  \begin{tabular}{lrr}
    \toprule
    Model             & BSER (\%) & Payload (bpp) \\
    \midrule
    Min-LSM-1         & 1.06\%    & 1.29          \\
    Min-LSM-2         & 0.67\%    & 0.42          \\
    Mean-LSM-1        & 2.22\%    & 3.89          \\
    Mean-LSM-2        & 3.14\%    & 2.21          \\
    Min-LSM-1-PS-0.6  & 0.74\%    & 0.80          \\
    Min-LSM-1-PS-0.8  & 0.66\%    & 0.80          \\
    Mean-LSM-1-PS-1.2 & 0.82\%    & 1.20          \\
    Mean-LSM-2-PS-1.2 & 0.93\%    & 1.20          \\
    \bottomrule
  \end{tabular}
  \caption{Different Embedding Strategies Comparison}%
  \label{tab:diff_strategies_comparison}
  \vspace{\baselineskip}
  In the model name part, the value after LSM is the number of bits masked during embedding process and the value after PS is the maximum payload capacity the embedded image is limited to during permutative straddling.
\end{table}

\figureSteganographyResultMeanLSMOne%

\subsection{Steganalysis Experiments}

To ensure that our model is robust to steganalysis methods, we test our models using StegExpose~\cite{StegExpose} with linear interpolation of detection threshold from 0.00 to 1.00 with 0.01 as the step interval. The ROC curve is shown in Figure~\ref{fig:roc_curves} where true positive stands for an embedded image correctly identified that there are hidden data inside while false positive means that a clean figure is falsely classified as an embedded image. The figure shows a comparison among our several models, StegNet~\cite{StegNet} and Baluja-2017~\cite{HIPS} plotted in dash-line-connected scatter data. It demonstrates that StegExpose can only work a little better than random guessing and most BASN models perform better than StegNet and Baluja-2017.

Our model is also further examined with learning-based steganalysis methods~\cite{SPAM,SRM,Yedroudj}. All of these models are trained with our cover and embedded images.Their corresponding ROC curves are shown in Figure~\ref{fig:roc_curves}. SRM~\cite{SRM} method works quite well on our model with a larger payload capacity, however in real-world applications we can always keep our dataset private and thus ensuring high security in resisting detection from learning-based steganalysis methods.

\figureROCCurves%

\subsection{Feature Distortion Analysis}

Figure~\ref{fig:basn_feature_distortion_rate} shows that our model has very little influence on targeted neural-network-automated tasks, which in this case is classification. Most embedded images, even carrying with more than 3 bpp of hidden information, takes an average of only 2\% distortion.

\figureBASNFeatureDistortionRate%

\section{Conclusion}

This paper proposes an image stagnography method based on a binary attention mechanism to ensure little influence steganography is made to neural-network-automated tasks. The first attention mechanism, image texture complexity (ITC) model, help track down the pixel locations and their tolerance of modification without being noticed by the human visual system. The second mechanism, minimizing feature distortion (MFD) model, further keeps down the embedding impact through feature map reconstruction. Moreover, some attention fusion and finetune techniques are also proposed in this paper to improve security and hidden information extraction accuracy. The imperceptibility of secret information
by our method is proved such that the embedding images can effectively resist detection by several steganalysis algorithms.

\bibliographystyle{plain}
\bibliography{main}

\end{document}